\documentclass[runningheads]{llncs}
\usepackage[T1]{fontenc}
\usepackage{graphicx}
\usepackage{booktabs}
\usepackage{multirow}
\usepackage{amsmath}
\usepackage{tabularx}
\usepackage{array}
\usepackage{amssymb,amsfonts}
\begin{document}
\title{Closed-Loop Graph Algorithm Execution with Small Language Models: Step Accuracy and Rollout Reliability}
\titlerunning{Closed-Loop Graph Algorithm Execution with Small Language Models}
%
\author{
Michal Podstawski
}
    
\authorrunning{M. Podstawski}

\institute{
NASK National Research Institute, Warsaw, Poland\\
\url{https://nask.pl/}
\\
\email{michal.podstawski@nask.pl}
}
\maketitle              
\begin{abstract}
Small language models offer an efficient alternative to large-scale systems,
but their ability to execute structured algorithms over multiple dependent
decisions remains poorly understood. We study graph algorithm execution as a
closed-loop prediction problem in which a model repeatedly selects the next
action from the current graph and algorithmic state. Our evaluation framework
covers several classical graph procedures, multiple synthetic graph families,
and disjoint training, validation, and test partitions. It assesses both local
decision quality and global execution behaviour using step accuracy, exact
rollout accuracy, constraint validity, partial solution quality, prefix
survival, and intervention-based diagnostics. The results show that adaptation
can produce reliable policies for structural procedures such as traversal and
coloring, while weighted algorithms remain substantially more sensitive to
error accumulation. More broadly, the findings demonstrate that strong
next-step prediction does not necessarily translate into reliable autonomous
execution and motivate evaluating algorithmic language models through complete
closed-loop rollouts rather than isolated decisions.

\end{abstract}

\section{Introduction}
\label{sec:introduction}

Language models are increasingly used in settings where a prediction is not a
terminal answer but an action that changes the state of an external system.
In such settings, reliability depends on more than the quality of isolated
predictions.  An action changes the input observed at the next step, so even a
small local error can move the system away from the states encountered during
training and alter every subsequent decision.  Classical graph algorithms
provide a controlled setting in which to study this distinction: their state
transitions are explicit, their correct actions can be generated
automatically, and their final outputs can be checked without a learned judge.


This paper studies small language models (SLMs) as closed-loop policies for
graph algorithm execution.  At each step, the model receives a textual
description of the graph and current symbolic state and emits one compact,
machine-parseable action.  A deterministic executor applies that action and
constructs the next state.  This separation keeps state transition semantics
outside the model while exposing the model to the consequences of its own
decisions.  We evaluate two instruction-tuned SLMs on six graph procedures and
compare teacher-forced next-action accuracy with autonomous rollout outcomes.

The experiments reveal a consistent separation between local competence and
global reliability.  Traversal policies retain their high step accuracy over
complete rollouts, and coloring remains comparatively robust.  In contrast,
weighted procedures exhibit large step-to-trajectory gaps: the models often
select plausible individual actions but fail to preserve correctness across a
complete execution.  Prefix-survival and intervention diagnostics localize
this failure and show that a single aggregate step score can obscure markedly
different execution behaviours.

The contributions of this work are:
\begin{itemize}
    \item a closed-loop formulation in which an SLM predicts one graph
          algorithm action and a symbolic executor owns the state transition;
    \item a controlled evaluation across six algorithms, three random graph
          families, and two independently adapted SLM architectures;
    \item a joint analysis of teacher-forced step accuracy, autonomous rollout
          accuracy, prefix survival, task-specific partial quality, and oracle
          correction requirements;
    \item empirical evidence that high next-action accuracy can coexist with
          severe rollout unreliability, particularly for weighted graph
          procedures.
\end{itemize}

\section{Related Work}
\label{sec:related_work}

Neural algorithmic reasoning investigates whether learned models can reproduce
the transitions of an algorithm rather than only predict its final output.
Early recurrent approaches learned short execution traces
\cite{zaremba2014learning}, while graph-based neural executors introduced
architectures aligned with relational algorithmic structure
\cite{velickovic2020neural}. The CLRS benchmark later provided standardized
intermediate supervision across many algorithms
\cite{velickovic2022clrs}, supporting research on multi-task transfer and
out-of-distribution generalization
\cite{ibarz2022generalist,bevilacqua2023causal}. This line of work primarily
studies learned processor design and algorithmic generalization; we instead use
graph algorithms to examine the execution reliability of autoregressive
language-model policies.

Recent work represents algorithmic execution as text. CLRS-Text trains
language models to generate textual traces and outputs
\cite{markeeva2024clrstext}, while MAGMA evaluates graph-algorithm reasoning at
intermediate execution stages \cite{taylor2024magma}. These studies focus on
generating or evaluating textual traces and intermediate states. Our formulation
instead treats the model as an action policy: it emits one compact action, and
an external symbolic executor applies the corresponding state transition. This
permits the same policy to be evaluated both on reference states and under
states induced by its own previous actions. The resulting comparison, together
with prefix-survival and oracle-correction diagnostics, directly measures how
local prediction errors accumulate during autonomous execution.

\section{Problem Formulation}
\label{sec:formulation}

Let $G=(V,E,w)$ be a graph, where $w$ is omitted for unweighted tasks.  For an
algorithm $A$, let $s_t$ denote its symbolic state after $t$ transitions and
$a_t$ a discrete action.  A deterministic transition function
\begin{equation}
    s_{t+1}=T_A(G,s_t,a_t)
\end{equation}
updates the state, and a termination predicate $D_A(G,s_t)$ indicates whether
execution is complete.  A deterministic reference implementation defines a
canonical action $a_t^\star$ for each reachable reference state and produces
a trajectory
\begin{equation}
    \tau^\star =
    \bigl((s_0^\star,a_0^\star),\ldots,
          (s_{T-1}^\star,a_{T-1}^\star)\bigr).
\end{equation}

The language model represents a policy
\begin{equation}
    \pi_\theta(a_t\mid x(G,s_t)),
\end{equation}
where $x(G,s_t)$ is a textual serialization of the graph and current state.
The output space is deliberately compact: one line represents one action.
The executor parses the line, validates its syntax, and applies the resulting
action through $T_A$.

Two evaluation regimes follow from this formulation.  In teacher-forced step
evaluation, the model is queried only on reference states $s_t^\star$.  This
measures whether it can identify the correct local transition when the input
state is known to be correct.  In autonomous rollout evaluation, execution
starts from $s_0^\star$, but every subsequent state is produced by applying the
model's own action.  Consequently, after an error the model may encounter a
state outside the reference trajectory.  The difference between the two
regimes is the principal object of study.

\section{Experimental Protocol}
\label{sec:protocol}

\subsection{Graphs and data splits}

We use graphs from the TinyGraphEstimator collection
\cite{podstawski2025tinygraph}.  The experimental subset contains 1,200
undirected graphs divided into 900 training, 150 validation, and 150 test
graphs.  Each split is balanced across Barab\'asi-Albert,
Erd\H{o}s-R\'enyi, and Watts-Strogatz generators: the training split contains
300 graphs from each family, while validation and test each contain 50 per
family.  Graphs contain between 20 and 30 vertices.  Split integrity is checked
using generator family, seed, graph size, and generator parameters; no such
provenance signature occurs in more than one split.

The source data are unweighted.  For Prim, Bor\r{u}vka, and Dijkstra, each
undirected edge receives a deterministic integer weight in $[1,10]$.  The
weight is obtained by hashing the graph provenance and canonicalized endpoint
pair, making weights stable across machines and repeated runs without exposing
test information during training.

\subsection{Algorithmic policies}

We evaluate BFS, DFS, greedy coloring, Bor\r{u}vka's minimum spanning tree
procedure, Dijkstra's shortest-path algorithm, and Prim's minimum spanning
tree procedure.  Source-based algorithms start from vertex~0.  A reference
implementation exposes a uniform interface for state initialization, canonical
next-action generation, action application, termination, and final-answer
extraction.  Table~\ref{tab:task_characteristics} summarizes the compact action
spaces and the amount of generated supervision.

Each training example contains the complete textual state followed by a single
target action.  BFS and DFS predict which newly discovered neighbours are
enqueued or pushed.  Coloring assigns a color to the current vertex.  Prim and
Bor\r{u}vka select one weighted edge, while Dijkstra selects the next visited
vertex together with its relaxations.  Prompts expose only information
available in the current algorithmic state; precomputed future actions are not
included.

\begin{table*}[t]
\centering
\caption{Task formulation and generated supervision. Pair counts report
training/validation next-action examples generated from 900/150 graphs;
mean length is calculated over 150 test trajectories.}
\label{tab:task_characteristics}
\small
\renewcommand{\arraystretch}{1.12}
\setlength{\tabcolsep}{4pt}

\begin{tabularx}{\textwidth}{
    @{}
    l
    >{\raggedright\arraybackslash}X
    r
    r
    l
    @{}
}
\toprule
Algorithm & Step action & Train/valid & Mean steps & Soft metric \\
\midrule
BFS
& Enqueue neighbours
& 22{,}586 / 3{,}736
& 25.47
& Order accuracy \\
DFS
& Push neighbours
& 22{,}586 / 3{,}736
& 25.47
& Order accuracy \\
Coloring
& Assign node color
& 22{,}586 / 3{,}736
& 25.47
& Color accuracy \\
Bor\r{u}vka
& Select component edge
& 21{,}686 / 3{,}586
& 24.47
& Edge F1 \\
Dijkstra
& Visit node and relax distances
& 22{,}586 / 3{,}736
& 25.47
& Distance accuracy \\
Prim
& Select cut edge
& 21{,}686 / 3{,}586
& 24.47
& Edge F1 \\
\bottomrule
\end{tabularx}
\end{table*}

\subsection{Models and adaptation}

We adapt Llama-3.2-1B-Instruct and Qwen2.5-1.5B-Instruct
\cite{dubey2024llama,qwen2024qwen25}.  A separate adapter is trained for every
model-algorithm pair.  The base weights are quantized to 4-bit NF4 with double
quantization, and LoRA \cite{hu2022lora,dettmers2023qlora} updates are applied to the query, key, value, output,
gate, up, and down projection modules.  We use rank 16, scaling factor 32, and
dropout 0.05.

Training minimizes causal language-model loss only on target action tokens;
prompt tokens are masked.  We use AdamW with learning rate $2\times10^{-4}$,
zero weight decay, a cosine schedule, 3\% warm-up, gradient clipping at 1.0,
and an effective batch size of 16 obtained from a physical batch size of one
and 16 gradient-accumulation steps.  Training runs for at most eight epochs
with seed 42.  The adapter with the lowest validation loss is retained, and
training stops after two non-improving epochs using a minimum improvement of
$10^{-4}$.

Maximum context length is selected by task: 1,024 tokens for BFS and DFS,
1,536 for coloring, and 2,048 for the weighted algorithms.  Evaluation uses
greedy decoding with no sampling or beam search and generates at most 64 new
tokens.  Only the first non-empty output line is parsed as the action.  All
reported test results use the validation-selected adapter and the fixed
150-graph test split.

\subsection{Unadapted baseline}

To separate adaptation from instruction following, we evaluate the unadapted
instruction models using the same graph states, action formats, deterministic
decoding, and 150 test graphs per algorithm. The zero-shot prompt specifies
the required one-line syntax but provides no demonstrations or parameter
updates. We additionally evaluate a chain-of-thought (CoT) variant that requests
a short rationale followed by the final executable action. Across the two
models and six algorithms, zero-shot prompting produces no exact trajectories
($0/1{,}800$ rollouts; $0.00\%$), and CoT prompting yields the same result
($0/1{,}800$; $0.00\%$). Although both settings occasionally predict an
individual action correctly, neither achieves a complete canonical execution
in any model-algorithm combination. Thus, the unadapted models exhibit
effectively no closed-loop execution ability in this setting, and eliciting
additional reasoning does not remedy the failure.

\section{Evaluation Measures}
\label{sec:metrics}

\paragraph{Step accuracy.}
Step accuracy is the fraction of reference states for which the parsed model
action exactly matches the canonical next action.  It is teacher-forced: every
query uses a correct reference state even if the model would have failed at an
earlier step.

\paragraph{Exact rollout accuracy.}
Beginning from the algorithm's initialized state, the model selects actions autonomously until termination or failure.  Exact rollout accuracy is the fraction of test graphs for which
the final answer equals the canonical reference answer.  Equality is defined
over the ordered traversal for BFS and DFS, the complete node assignment for
coloring and Dijkstra, and the canonicalized selected-edge set for Prim and
Bor\r{u}vka.  This measure therefore evaluates the result of a complete
closed-loop execution, not merely the final teacher-forced step.

\paragraph{Constraint validity.}
We additionally report a deliberately weaker, task-specific constraint check.
For traversal it verifies a source-rooted, duplicate-free sequence containing
only reachable vertices; for Prim and Bor\r{u}vka it verifies that selected
edges form a spanning tree; for Dijkstra it verifies a complete non-negative
distance labeling satisfying edge-wise inequalities; and for coloring it
verifies a complete conflict-free assignment.  This diagnostic should not be
interpreted as exact task accuracy: for example, it does not test minimum
weight for a spanning tree and does not require a traversal sequence to contain
every reachable vertex.

\paragraph{Soft score.}
The soft score measures partial agreement with the reference answer.  It is
order-position accuracy for BFS and DFS, node-color accuracy for coloring,
edge F1 for Prim and Bor\r{u}vka, and exact node-distance accuracy for
Dijkstra.  Because these definitions differ, soft scores are comparable
between models within an algorithm but not across algorithms.

\paragraph{Rollout diagnostics.}
The step-to-trajectory gap is the difference between step and exact rollout
accuracy.  Prefix survival at position $k$ is the fraction of test rollouts
whose first $k$ actions match the reference trajectory; prefix AUC is the mean
of this survival curve across eligible positions.  The first-error index is
the first position at which an erroneous rollout deviates from the reference.
Finally, an intervention rollout queries the model at every reference-corrected
state, advances using the oracle action, and counts how many model predictions
would require replacement.  The reported correction count is the mean number
of such replacements per graph.

\section{Results}
\label{sec:results}


The unadapted models produce no exact autonomous rollout on any of the six
algorithms.  This result is not explained solely by malformed output: the
models occasionally select correct individual actions, but those actions do
not form a complete canonical execution.  Parameter-efficient adaptation
changes this behaviour substantially.

Table~\ref{tab:main_results} reports the main results.  BFS and DFS are the
most reliable tasks.  Llama reaches 97.33\% and 96.00\% exact rollout accuracy,
respectively, while Qwen reaches 95.33\% and 94.00\%.  Their step accuracies
are also high, but the more important observation is that this local
competence survives a complete autonomous execution.  Coloring is moderately
harder: rollout accuracy is 80.00\% for Llama and 86.00\% for Qwen despite
step accuracy above 98\% for both models.

\begin{table*}[t]
\centering
\caption{Performance on 150 held-out test graphs. All values are percentages.
Step denotes next-action accuracy, Traj.\ exact rollout accuracy, Valid
task-specific constraint validity, and Soft partial agreement with the
canonical reference solution. Validity and soft-score definitions differ by
algorithm and are specified in Section~\ref{sec:metrics}.}
\label{tab:main_results}
\small
\renewcommand{\arraystretch}{1.1}

\begin{tabular*}{\textwidth}{
    @{\extracolsep{\fill}}
    l rrrr rrrr
    @{}
}
\toprule
& \multicolumn{4}{c}{Llama-3.2-1B-Instruct}
& \multicolumn{4}{c}{Qwen2.5-1.5B-Instruct} \\
\cmidrule(lr){2-5}
\cmidrule(lr){6-9}
Algorithm
& Step & Traj. & Valid & Soft
& Step & Traj. & Valid & Soft \\
\midrule
BFS
& 99.90 & 97.33 & 100.00 & 99.69
& 96.99 & 95.33 & 99.33 & 97.24 \\
DFS
& 99.84 & 96.00 & 98.00 & 98.91
& 96.96 & 94.00 & 100.00 & 96.03 \\
Coloring
& 98.77 & 80.00 & 89.33 & 98.43
& 99.24 & 86.00 & 93.33 & 98.28 \\
Bor\r{u}vka
& 92.70 & 21.33 & 54.67 & 95.01
& 92.92 & 25.33 & 54.67 & 95.85 \\
Dijkstra
& 83.54 & 13.33 & 13.33 & 86.17
& 86.68 & 17.33 & 18.00 & 83.69 \\
Prim
& 93.90 & 41.33 & 62.67 & 95.97
& 87.74 & 14.67 & 39.33 & 90.60 \\
\bottomrule
\end{tabular*}
\end{table*}


The weighted procedures display a different pattern.  Bor\r{u}vka step
accuracy is approximately 93\% for both models, but exact rollout accuracy is
only 21.33\% for Llama and 25.33\% for Qwen.  Dijkstra yields 83.54-86.68\%
step accuracy but only 13.33-17.33\% exact rollouts.  Prim shows the largest
between-model difference: Llama reaches 41.33\% exact rollout accuracy, whereas
Qwen reaches 14.67\%.

Task-specific soft scores remain high for the minimum-spanning-tree procedures
even when exact rollout accuracy is low.  This indicates that failed rollouts
often retain many reference edges, not that they are exact or optimal
solutions.  The constraint-validity column should be read with the same care:
it measures structural feasibility under the checks in
Section~\ref{sec:metrics}, not equivalence to the reference algorithm.


Table~\ref{tab:rollout_diagnostics} explains why similar step scores can lead
to very different execution outcomes.  Traversal has a small
step-to-trajectory gap and prefix AUC above 93\% for both models.  Coloring
retains a high prefix AUC but accumulates enough late errors to reduce exact
rollout accuracy.  Bor\r{u}vka exhibits gaps above 67 percentage points even
though its prefix AUC remains around 70\%, suggesting that many rollouts begin
correctly but do not preserve the sequence to completion.

Dijkstra fails earlier and more persistently.  Its prefix AUC is only
15.47\% for Llama and 18.28\% for Qwen, and completing a reference-corrected
trajectory requires on average 4.19 and 3.39 interventions, respectively.
Prim again reveals a model-specific difference: Llama has a smaller gap,
higher prefix AUC, and fewer required corrections than Qwen.  These diagnostics
show that the overall hierarchy is not an artefact of any single accuracy
definition.

\begin{table*}[t]
\centering
\caption{Closed-loop rollout diagnostics on 150 held-out test graphs. Gap is
the difference between step and exact rollout accuracy in percentage points.
AUC is correct-prefix survival summarized over the rollout and reported as a
percentage. Err.\ is the mean one-based position of the first incorrect action
among erroneous rollouts. Corr.\ is the mean number of oracle corrections
required per rollout.}
\label{tab:rollout_diagnostics}
\small
\renewcommand{\arraystretch}{1.12}
\setlength{\tabcolsep}{3pt}

\begin{tabularx}{\textwidth}{
    @{}
    l
    *{8}{>{\centering\arraybackslash}X}
    @{}
}
\toprule
& \multicolumn{4}{c}{Llama-3.2-1B-Instruct}
& \multicolumn{4}{c}{Qwen2.5-1.5B-Instruct} \\
\cmidrule(lr){2-5}
\cmidrule(lr){6-9}
Algorithm
& Gap & AUC & Err. & Corr.
& Gap & AUC & Err. & Corr. \\
\midrule
BFS
& 2.57 & 97.78 & 10.25 & 0.03
& 1.66 & 94.30 & 6.63 & 0.77 \\
DFS
& 3.84 & 96.95 & 17.00 & 0.04
& 2.96 & 93.85 & 6.89 & 0.77 \\
Coloring
& 18.77 & 92.29 & 23.57 & 0.31
& 13.24 & 93.85 & 21.67 & 0.19 \\
Bor\r{u}vka
& 71.37 & 73.10 & 20.74 & 1.79
& 67.58 & 70.43 & 20.03 & 1.73 \\
Dijkstra
& 70.21 & 15.47 & 4.46 & 4.19
& 69.35 & 18.28 & 4.29 & 3.39 \\
Prim
& 52.56 & 61.36 & 15.32 & 1.49
& 73.08 & 42.43 & 12.28 & 3.00 \\
\bottomrule
\end{tabularx}
\end{table*}

\section{Discussion}
\label{sec:discussion}

The main empirical conclusion is that next-action accuracy and autonomous
execution answer different questions.  Step accuracy measures performance on
states produced by the reference algorithm.  A rollout additionally tests
whether the model remains correct after repeated application and whether it
can handle states induced by its own earlier decisions.  For BFS and DFS,
these quantities are closely aligned.  For the weighted procedures, they are
not: an apparently strong local classifier can still be an unreliable policy.

The two evaluation regimes also aggregate errors differently. Step accuracy
micro-averages decisions over all reference states, whereas exact rollout
accuracy counts a graph as correct only when its complete execution matches the
reference. Errors concentrated in a small number of difficult graphs may
therefore reduce step accuracy while leaving most rollouts correct, as observed
for Qwen on BFS. Conversely, errors distributed across many graphs can sharply
reduce rollout accuracy even when the aggregate step score remains high, as in
Bor\r{u}vka. Some decline is also expected without feedback effects because a
trajectory succeeds only if every required action is correct. The
step-to-trajectory gap alone should therefore not be interpreted as proof of
degradation under self-induced states. Prefix survival, first-error position,
and correction counts provide the additional evidence needed to distinguish
late, concentrated errors from early or persistent rollout failure.

This distinction matters beyond graph algorithms.  Language-model agents,
tool-using systems, and iterative planners also alter their future inputs by
acting.  Evaluations based only on isolated decisions may therefore
overestimate operational reliability.  The deterministic graph environment
used here makes this effect measurable without confounding it with stochastic
tools or subjective outcome grading.

The results suggest that output length alone is not the main source of
difficulty: the mean trajectory lengths differ little across the tasks.
Weighted algorithms instead require repeated comparison of numerical edge or
distance values while preserving globally coupled state.  A locally plausible
edge selection or relaxation can alter the frontier, component structure, or
distance map used by every later decision.  Coloring also requires sequential
state maintenance, but its constraint is local and errors tend to occur later,
as reflected by its high prefix AUC.

The difference between Prim and Bor\r{u}vka further indicates that ``weighted''
is not a complete explanation.  The action representation, state
serialization, tie-breaking policy, and way an error changes subsequent state
all influence rollout reliability.  These factors should be separated in
future controlled ablations rather than compressed into a single notion of
algorithm difficulty.

\section{Limitations}
\label{sec:limitations}

The study has several limitations.  First, all graphs are synthetic and
contain only 20-30 vertices.  The test split changes graph instances but not
the generator families or size range, so the results establish
within-distribution generalization rather than extrapolation to larger or
real-world graphs. Second, weights are deterministic synthetic integers;
different weight distributions or tie frequencies may change the relative
difficulty of the weighted tasks. Third, exact rollout accuracy follows one deterministic reference
implementation and tie-breaking convention.  Alternative outputs can be
acceptable for some tasks, but the reported constraint-validity diagnostic is
intentionally weak and heterogeneous across algorithms.  It should not be
treated as a substitute for exact correctness.

\section{Conclusion}
\label{sec:conclusion}

We evaluated small language models as closed-loop policies for six classical
graph algorithms.  Parameter-efficient adaptation produced reliable traversal
and coloring policies, but weighted procedures remained fragile despite high
next-action accuracy. Prefix-survival and correction analyses showed how local errors are distributed
and accumulate across autonomous execution.

The broader implication is methodological: evaluating an algorithmic language
model only on isolated reference states can substantially overstate its
operational reliability.  When predictions change future inputs, complete
closed-loop rollouts should be treated as a primary evaluation unit.  Future
work should test larger and out-of-distribution graphs, multiple training
seeds, shared multi-algorithm adapters, and stronger semantic validity checks.

\begin{credits}
\subsubsection{\ackname} This manuscript acknowledges the use of ChatGPT~\cite{openai2026chatgpt}, powered by the GPT-5.5 language model developed by OpenAI, to improve language clarity, refine sentence structure, and enhance overall writing precision.
\end{credits}

\end{document}